\UseRawInputEncoding
\documentclass[letterpaper, 10 pt, conference]{ieeeconf}  

\usepackage{booktabs}
\usepackage{amssymb}
\usepackage{amsmath}
\usepackage{graphicx}
\usepackage{tabularx}
\usepackage{array}
\usepackage{algorithm}
\usepackage{algpseudocode}
\usepackage[table]{xcolor}
\usepackage[colorlinks=true, urlcolor=blue]{hyperref}

\IEEEoverridecommandlockouts                              

\title{\LARGE \bf

Talk, Render, Act: Integrating Social Gesture  and Digital Face with Synchronized Speech for Conversational Humanoid Robot

}

\author{
Jin Jiang\textsuperscript{1} \quad
Kun Li\textsuperscript{1} \quad
Jiancong Ma\textsuperscript{1} \quad
Shengcai Liao\textsuperscript{1$\dag$}
\\[2mm]
\textsuperscript{1}United Arab Emirates University (UAEU)
         \thanks{$^{1}$Jin Jiang, Kun Li, Jiancong Ma, and Shengcai Liao are with the College of Computing and Artificial Intelligence (CCAI), United Arab Emirates University (UAEU), Al Ain, Abu Dhabi, United Arab Emirates.}
         \thanks{Corresponding author: Shengcai Liao (email: scliao@ieee.org).}
 }

\usepackage{graphicx}

\usepackage{makecell}

\begin{document}

\maketitle
\thispagestyle{empty}
\pagestyle{empty}

\begin{abstract}
Expressive humanoid interaction requires speech, facial animation, and body gestures to form a coherent response. However, many full-body humanoid robots produce speech and gestures without a visually expressive face, while talking-face animation and robot gesture generation are typically developed separately. We present Talk, Render, Act (TRABot), an agent-based framework comprising specialized agents for motion-atom construction, dialogue generation, motion planning, and facial animation. First, to produce natural and semantically meaningful gestures, we construct Robot-Ready Semantic Motion Atoms by segmenting long-form, G1-retargeted BEAT2 motion into units with natural gesture boundaries, human-verified communicative functions, and feasible trajectories. Second, to preserve semantic order and coordinate body motion with the spoken response, we introduce a Semantic-Conditioned Compositional Planner. Given an ordered semantic function sequence and an estimated response duration, the planner selects approved atoms to realize the longest feasible action sequence while accounting for transitions and neutral recovery. Finally, we deploy a Streaming Face-Speech-Body Integration system on a physical G1 humanoid, combining streaming dialogue audio, audio-driven facial animation, and semantically planned body motion in a unified real-time interaction loop. Quantitative and qualitative experiments demonstrate that TRAbot achieves the best overall performance among all compared conditions in terms of naturalness, expressiveness, and multimodal coherence. Our project page is available at
\href{https://trabot-humanoid.github.io/}{https://trabot-humanoid.github.io/}.
\end{abstract}

\section{Introduction}
\label{sec:introduction}

Human face-to-face communication relies on the coordination of verbal and nonverbal behaviors. Speech conveys linguistic content, whereas facial movements and body gestures provide complementary cues for affective expression, emphasis, and deictic reference \cite{cassell2001beat, kucherenko2020gesticulator,CommunicativeRobotGesture2012}. Consequently, expressive humanoid interaction requires speech, facial animation, and body motion to be coordinated semantically and temporally \cite{wang2024misbehavin,hogue2025joint,yoon2019robots}.

Recent advances have substantially improved the individual components of embodied interaction. Speech-driven talking-face models generate articulatory and expressive facial motion from audio \cite{yu2026soulx,huang2025liveavatar,fan2022faceformer,richard2021meshtalk,xing2023codetalker}. Social robots such as Furhat and Haru combine dialogue with expressive facial displays, gaze, or platform-specific physical behaviors \cite{cherakara2023furchat,gomez2020holistic,wang2024misbehavin}. These systems demonstrate the value of facial embodiment, but they do not provide the expressive motion of a full-body humanoid. Conversely, co-speech gesture methods generate semantically or rhythmically appropriate body motion for virtual agents and robots \cite{kucherenko2020gesticulator,yoon2019robots}. Recent work further enables streaming, semantically aligned gestures on physical humanoids \cite{wang2026robogesture}, but focuses on speech and body motion without a continuously animated talking face. Holistic animation methods jointly model facial and bodily behavior for digital humans, including real-time interactive generation \cite{chen2024diffsheg,hogue2025joint,Mughal2026CVPR}. Such systems operate in a shared virtual representation and do not address robot-specific motion feasibility or state-dependent execution. Taken together, these approaches leave the coordination of streaming speech, an audio-driven digital face, and semantically planned humanoid motion within a unified conversational system insufficiently explored.

A direct combination of existing components does not fully resolve this problem because facial and bodily behaviors involve different representations and time scales. Facial animation can be generated frame by frame from streaming audio. Humanoid body motion must instead be organized into semantically meaningful actions, validated for the target embodiment, connected from the current robot state, and composed within the duration of a spoken response. Dialogue generation operates at the level of communicative intent and reply duration, whereas motion execution requires concrete trajectories that satisfy these constraints. These differences motivate an agent-based architecture in which each agent performs a distinct generation or decision process and communicates through explicit semantic and temporal variables.

We present Talk, Render, Act (TRABot), an agent-based framework comprising one offline agent and three online agents. The offline Motion Atom Construction Agent transforms long-form, G1-retargeted BEAT2 motion \cite{liuhaiyang2024emage} into Robot-Ready Semantic Motion Atoms with natural gesture boundaries, human-verified communicative functions, and feasible trajectories. The resulting atoms provide inspectable and reusable robot actions while establishing a semantic interface between dialogue generation and motion planning.

During interaction, the Dialogue Agent generates a spoken reply together with an ordered function sequence and an estimated response duration. The Motion Planning Agent implements our Semantic-Conditioned Compositional Planner, which selects approved atoms to realize the longest feasible action sequence within the available response duration while accounting for transitions and neutral recovery. A state-aware executor recomputes each transition from the robot state reached during execution. In parallel, the Facial Animation Agent generates facial motion from the same streaming audio used for speech playback. The Streaming Face-Speech-Body Integration system executes the resulting speech, facial animation, and body motion concurrently within a unified real-time interaction loop. This organization preserves the modality-specific processing required by facial rendering and robot execution while coordinating all three modalities through a shared spoken response.

The main contributions of this work are threefold.
\begin{itemize}
\item We construct \textbf{Robot-Ready Semantic Motion Atoms} by segmenting long-form, G1-retargeted BEAT2 motion into units with natural gesture boundaries, human-verified communicative functions, and feasible trajectories.

\item We introduce a \textbf{Semantic-Conditioned Compositional Planner} that maps an ordered function sequence and a response-duration estimate to the longest feasible action sequence while accounting for transitions, neutral recovery, and the current robot state.

\item We develop a \textbf{Streaming Face-Speech-Body Integration System} that coordinates dialogue audio, audio-driven facial animation, and semantically planned humanoid motion through specialized agents in a unified real-time interaction loop.

\end{itemize}

\section{Related Work}

\subsection{Co-Speech Gesture Generation for Embodied Agents}

Co-speech gesture generation has progressed from manually specified behavior rules to data-driven motion synthesis~\cite{ConvoFusion2024,HOP2025,GestureHYDRA2025,SemGes2025}. Yoon et al. learned a mapping from speech transcripts to gesture sequences and demonstrated real-time execution on a NAO robot~\cite{yoon2019robots}. Gesticulator subsequently combined semantic and acoustic speech representations to generate joint rotations that capture both content-related and rhythmic gestures~\cite{kucherenko2020gesticulator}. Other approaches incorporate speaker identity and multimodal speech context to model individual gesturing styles~\cite{yoon2020trimodal}, or represent motion using discrete gesture tokens to better capture the one-to-many relationship between speech and movement~\cite{lu2023cospeech}. Context-aware systems have further conditioned robot gestures on nonverbal cues from an interaction partner~\cite{nguyen2023gesticulating}. Although these methods improve the appropriateness, diversity, and responsiveness of generated motion, they primarily focus on the articulated body and commonly treat facial behavior as a separate component.

\subsection{Facial and Holistic Animation}

Speech-driven facial animation has traditionally focused on accurate articulation and plausible full-face motion. MeshTalk separates audio-correlated articulation from weakly correlated behavior such as blinks and eyebrow movements~\cite{richard2021meshtalk}, while FaceFormer exploits long-range speech context through autoregressive attention~\cite{fan2022faceformer}. CodeTalker instead maps speech to a learned discrete motion prior to alleviate the over-smoothing associated with direct motion regression~\cite{xing2023codetalker}.
Recent work has further advanced speech-driven facial animation toward greater photorealism, expressiveness, and real-time performance. SyncTalk combines explicit synchronization control with neural radiance-field rendering to improve the alignment of speech, lip motion, facial expression, and head pose \cite{peng2024synctalk}. EMO adopts direct audio-to-video diffusion without relying on intermediate 3D models or facial landmarks \cite{tian2024emo}, while FLOAT performs flow matching in a learned motion latent space for expressive and temporally consistent portrait animation \cite{ki2025float}. Teller supports real-time streaming generation through autoregressive motion modeling \cite{zhen2025teller}. SoulX-FlashHead and LiveAvatar further enable long-duration facial animation from streaming audio \cite{yu2026soulx,huang2025liveavatar}. 
Although these methods produce high-quality facial motion, they do not address its coordination with actions executed by a separately embodied physical robot.

\subsection{Multimodal Expressive Behavior for Social Robots}

The coordination of verbal and nonverbal behavior has a long history in embodied-agent systems. BEAT transforms linguistic input into synchronized speech, gaze, facial displays, and gestures using explicit behavioral rules~\cite{cassell2001beat}. For hybrid physical-digital embodiments, Gomez et al. coordinate Haru's mechanical motion and digital displays through animator-authored multimodal expressions~\cite{gomez2020holistic}. More recent systems use language models to reduce manual behavior authoring. FurChat combines LLM-generated dialogue with facial expressions on the Furhat robot~\cite{cherakara2023furchat}, while Expressive Furhat generates context-dependent facial gestures and gaze behavior in real time~\cite{abbo2026expressive}. On Haru, Wang et al. use conversational and affective cues to select expressive speech styles and pre-authored physical behaviors~\cite{wang2024misbehavin}.

LLMs have also been used to generate robot behavior more directly. GenEM translates natural-language descriptions of social behavior into parameterized control programs composed from available robot skills~\cite{mahadevan2024genem}. Galatolo and Winkle attach robot-specific behavior heads to a language model, enabling simultaneous generation of text and high-level communicative intentions that are mapped to facial expressions on Furhat or bodily gestures on Pepper~\cite{galatolo2025simultaneous}. Most closely related, Nichols et al. generate synchronized timelines for vocal, facial, and bodily expression on the Haru robot from a shared semantic analysis of dialogue~\cite{nichols2026holistic}. Building on these complementary directions, our work considers a hybrid embodiment in which robot actions and digital facial animation are generated as composable outputs of a shared agent. Our focus is on preserving semantic and temporal coherence between the two channels while satisfying the distinct execution requirements of physical actuation and facial rendering.
\section{Methodology}
\label{sec:method}

\subsection{TRABot Overview and Agent Interfaces}
\label{sec:overview}

\begin{figure*}[t]
\centering
\includegraphics[width=\textwidth]{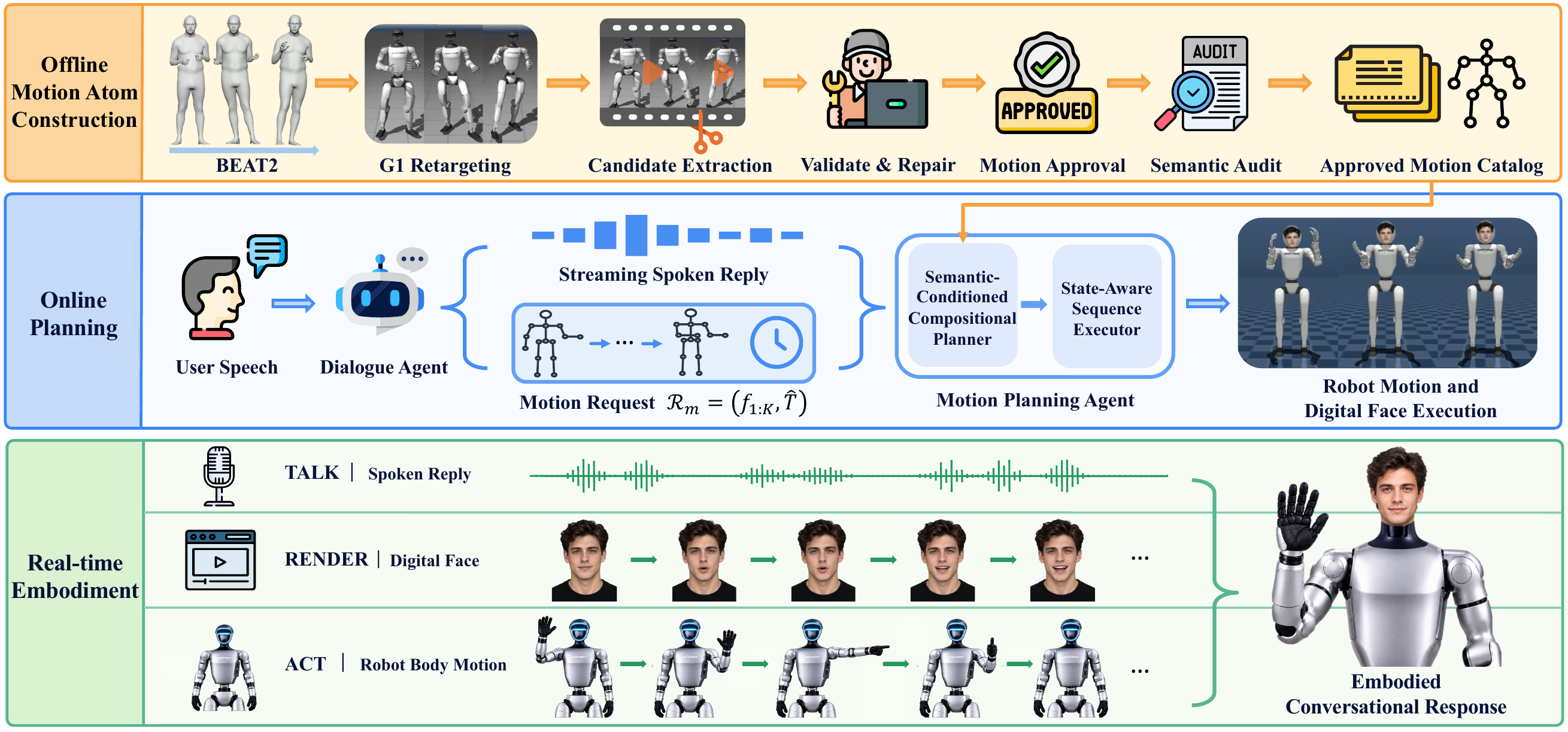}
\caption{Overview of TRABot. Offline Motion Atom Construction Agent converts G1-retargeted BEAT2 motion into an approved catalog of Robot-Ready Semantic Motion Atoms. 
Online the Dialogue Agent generates streaming reply audio and a semantic motion request. 
The Motion Planning Agent composes approved atoms and constructs state-dependent transitions, while the Facial Animation Agent generates facial frames from the same audio stream. 
The Streaming Face-Speech-Body Integration System executes speech, facial animation, and robot body motion concurrently.}
\label{fig:overview}
\end{figure*}

TRABot comprises one offline agent and three online agents connected through explicit semantic and temporal interfaces. The offline Motion Atom Construction Agent converts long-form, G1-retargeted BEAT2 motion into an approved catalog of Robot-Ready Semantic Motion Atoms. During interaction, the Dialogue Agent generates streaming reply audio together with a semantic motion request
\begin{equation}
\mathcal{R}_{m}=(f_{1:K},\hat T),
\label{eq:request}
\end{equation}
where $f_{1:K}$ is the ordered sequence of communicative functions produced by the dialogue model, and $\hat T$ is its estimated reply duration.

The local catalog
$\mathcal{C}=\{a_i\}_{i=1}^{N}$ contains only approved motion atoms. For online planning, each atom is represented as
\begin{equation}
a_i=(\phi_i,Q_i,d_i,h_i),
\label{eq:atom}
\end{equation}
where $\phi_i$ is its human-audited communicative function, $Q_i$ is the
executable joint trajectory, $d_i$ is its duration, and $h_i$ denotes its
handedness (left, right, or bilateral). The boundary poses used for transition planning are obtained
directly from the trajectory:
\begin{equation}
q_i^{\mathrm{in}}=Q_i(0),
\qquad
q_i^{\mathrm{out}}=Q_i(d_i).
\label{eq:atom_boundaries}
\end{equation}

Given $\mathcal{R}_{m}$, the measured robot state $q$, and the remaining motion budget $B(\hat T)$, the Motion Planning Agent applies the Semantic-Conditioned Compositional Planner to select the longest budget-feasible atom sequence. Its State-Aware Sequence Executor constructs transitions from the current state and replans the remaining functions after each executed atom. In parallel, the Facial Animation Agent converts the streaming reply audio into facial frames. The Streaming Face-Speech-Body Integration System presents the resulting speech, facial animation, and body motion concurrently. Fig.~\ref{fig:overview} summarizes the complete pipeline.

\subsection{Robot-Ready Semantic Motion Atoms}
\label{sec:atoms}

The offline Motion Atom Construction Agent converts long-form, G1-retargeted BEAT2 motion into a catalog of naturally bounded, feasible, and semantically verified motion atoms.

\subsubsection{Motion Representation and Preprocessing}

We use BEAT2 conversational motion retargeted to the 29-DoF G1 kinematic convention. The floating root remains at its reference pose, and the 14 arm joints form the controlled motion space. Sequences with inconsistent joint layouts or invalid numerical values are rejected before further processing.

To identify stable gesture boundaries while preserving the executable motion, we maintain separate representations for boundary detection and robot execution. A copy of each sequence is resampled to 25~Hz and filtered with a three-frame median filter followed by the symmetric binomial kernel $[1,4,6,4,1]/16$ for boundary detection, while the original 30~Hz trajectory is retained for constructing executable motion atoms.

\subsubsection{Adaptive extraction and natural boundaries}

The preprocessing stage provides two synchronized representations of each recording: a smoothed 25~Hz signal for detecting motion and the original 30~Hz trajectory for constructing executable atoms. We first use the detection signal to identify coarse regions of intentional movement.

Because motion magnitude varies substantially across speakers and recordings, the agent estimates activity relative to each recording rather than using a global velocity threshold. For the 14 controlled arm joints, we use equal weighting and normalize each joint velocity by its kinematic range $r_j=q_j^{\max}-q_j^{\min}$ obtained from the G1 model. 
For frame $t$, the resulting activity energy is
\begin{equation}
E_t=
\sqrt{\frac{1}{14}
\sum_{j=1}^{14}
\left(\frac{\dot q_{t,j}}{r_j}\right)^2}.
\label{eq:energy}
\end{equation}
This normalization prevents joints with larger ranges of motion from dominating the activity signal.

Rather than applying a fixed activity threshold to every recording, we estimate its motion floor, dynamic range, and background variation from its own energy distribution. Let $P_p(E)$ denote the $p$-th percentile of the frame-wise energy values $\{E_t\}_{t=1}^{T}$. We define
\begin{equation}
b=P_{20}(E),\qquad
\Delta=P_{95}(E)-b,
\end{equation}
where $b$ approximates the low-activity energy floor and $\Delta$ describes the usable motion range of the recording. To estimate background variation without being affected by high-energy gesture strokes, we retain only the samples below the median energy,
\begin{equation}
\mathcal{E}_{\mathrm{quiet}}
=\{E_t\mid E_t\leq P_{50}(E)\},
\end{equation}
and compute their robust scale as
\begin{equation}
\sigma
=1.4826\,
\operatorname{median}_{e\in\mathcal{E}_{\mathrm{quiet}}}
\left|
e-\operatorname{median}(\mathcal{E}_{\mathrm{quiet}})
\right|.
\end{equation}
Here, $\sigma$ measures the amount of residual motion or retargeting noise present during relatively quiet portions of the recording.

Using these recording-specific statistics, we define separate onset and offset thresholds:
\begin{equation}
\begin{aligned}
\theta_{\mathrm{on}}
&=b+\operatorname{clip}\!\left(
\max(\kappa_{\mathrm{on}}\sigma,\alpha_{\mathrm{on}}\Delta),
\alpha_{\mathrm{on}}\Delta,\beta_{\mathrm{on}}\Delta\right),\\
\theta_{\mathrm{off}}
&=b+\operatorname{clip}\!\left(
\max(\kappa_{\mathrm{off}}\sigma,\alpha_{\mathrm{off}}\Delta),
\alpha_{\mathrm{off}}\Delta,\beta_{\mathrm{off}}\Delta\right),
\end{aligned}
\label{eq:hysteresis}
\end{equation}
where $\operatorname{clip}(x,l,u)$ restricts $x$ to the interval $[l,u]$.
The coefficients $\kappa_{\mathrm{on}}$ and $\kappa_{\mathrm{off}}$ determine the sensitivity to the estimated background noise, while $\alpha_{\mathrm{on}}$ and $\alpha_{\mathrm{off}}$ define lower bounds relative to the recording's dynamic range. The coefficients $\beta_{\mathrm{on}}$ and $\beta_{\mathrm{off}}$ prevent the thresholds from becoming excessively high. These coefficients are fixed hyperparameters shared by all recordings, and only $b$, $\Delta$, and $\sigma$ adapt to the current sequence. The higher onset threshold requires a clear increase above both the background noise and the recording's motion floor before starting an Atom. 
The lower offset threshold allows an ongoing gesture to pass through a brief slowdown without being split into multiple segments. 

To further suppress short fluctuations, an interval becomes active only after $E_t$ remains above $\theta_{\mathrm{on}}$ for 0.12~s and ends after it remains below $\theta_{\mathrm{off}}$ for 0.16~s. Inactive gaps shorter than 0.12~s are merged. A displaced stationary hold remains within the gesture when a retraction follows within 1.0~s. Recordings with $\Delta<4\sigma$ are rejected as low-signal sequences.

Threshold crossings provide coarse activity intervals, but they do not necessarily coincide with natural gesture boundaries. In particular, directly cutting at a threshold crossing can remove part of the preparation or retraction phase. We therefore refine each provisional endpoint by searching up to 0.32~s backward from each provisional onset and up to 0.32~s forward from each provisional offset for a locally stable pose. 
Candidate boundaries are evaluated using
\begin{equation}
C_t^{\mathrm{bdry}}
=
\lambda_E\bar E_t+
\lambda_A\bar A_t+
\lambda_P\bar P_t,
\label{eq:boundary}
\end{equation}
where $\bar E_t$, $\bar A_t$, and $\bar P_t$ are normalized motion energy, acceleration energy, and local pose variation. $\lambda_E$, $\lambda_A$, and $\lambda_P$ are fixed boundary-cost weights shared across all recordings. Low values indicate that the robot is moving slowly, with little acceleration, near a stable pose. An endpoint with cost above $0.50$ is rejected. Intervals longer than $4$~s are split only at feasible low-cost valleys using dynamic programming with a duration penalty, and every resulting segment must have a duration between 0.4~s and 4~s. This refinement favors locally stable poses and avoids cutting through the stroke of a gesture. 

\begin{table}[t]
\centering
\caption{Planner-facing communicative vocabulary. Communicative functions are used as online planning labels, whereas common gesture families are inherited from the original BEAT2 annotations and retained as offline metadata.}
\label{tab:functions}
\resizebox{\linewidth}{!}{%
\begin{tabular}{ccc}
\toprule
Category & Communicative functions & Common family \\
\midrule
emphasis
& \texttt{emphasize}
& beat \\

enumeration
& \texttt{enumerate}
& beat / metaphoric \\

contrast
& \texttt{contrast\_compare}
& metaphoric \\

reference
& \makecell[c]{\texttt{refer\_self}, \texttt{refer\_other},\\
           \texttt{refer\_external}}
& deictic \\

depiction
& \makecell[c]{\texttt{depict\_size\_shape},\\
           \texttt{depict\_direction\_motion}}
& iconic \\

turn management
& \makecell[c]{\texttt{turn\_open},\\
           \texttt{turn\_yield}}
& \makecell[c]{beat / metaphoric /\\ emblem} \\

social
& \makecell[c]{\texttt{greet\_farewell}, \texttt{thank},\\
           \texttt{affirm}, \texttt{deny}}
& emblem \\

stance
& \makecell[c]{\texttt{certainty}, \texttt{uncertainty},\\
           \texttt{enthusiasm}}
& metaphoric / emblem \\
\bottomrule
\end{tabular}%
}
\end{table}

\subsubsection{Feasibility validation and repair}

A motion segment that appears plausible in the source data may become infeasible after retargeting because of joint-limit violations, excessive derivatives, or unexpected contacts. We therefore validate every candidate in the target G1 configuration before it becomes eligible for Motion Approval. Joint positions must remain at least $0.05$~rad inside the configured limits, while peak velocity, acceleration, and jerk are constrained to $7$~rad/s, $120$~rad/s$^2$, and $2200$~rad/s$^3$, respectively. During a fixed-root MuJoCo rollout, we additionally reject contacts that are absent from the neutral configuration or become more than $2$~mm deeper, excluding the expected foot and ankle contacts.

A candidate that passes this gate proceeds to Motion Approval, whereas a failed candidate enters a deterministic repair search. We generate a finite set of repair variants using combinations of limb-wise amplitude scaling, endpoint-preserving smoothing, and temporal stretching. Amplitude scaling reduces excessive joint excursions, smoothing suppresses local trajectory fluctuations, and temporal stretching addresses the remaining dynamic violations. 

For each spatially repaired variant, we compute the safety-adjusted temporal stretch required by its derivative peaks. Let $v_j^{\max}$, $a_j^{\max}$, and $j_j^{\max}$ denote the maximum velocity, acceleration, and jerk of joint $j$, and let $v_j^{\mathrm{lim}}$, $a_j^{\mathrm{lim}}$, and $j_j^{\mathrm{lim}}$ denote their corresponding limits. The stretch factor is
\begin{equation}
s = 1.02\max_j\left\{ 1,\,
  \frac{v_j^{\max}}{v_j^{\mathrm{lim}}},\,
  \left(\frac{a_j^{\max}}{a_j^{\mathrm{lim}}}\right)^{1/2},\,
  \left(\frac{j_j^{\max}}{j_j^{\mathrm{lim}}}\right)^{1/3}
\right\}.
\label{eq:retime}
\end{equation}
Uniform stretching by $s$ reduces velocity, acceleration, and jerk by factors of $s$, $s^2$, and $s^3$, respectively, while $1.02$ provides a small numerical margin.

Every repaired variant is passed through the complete validation procedure again. Among the feasible alternatives, we retain the variant that best preserves the original trajectory shape, apex, endpoints, and duration. Recipes requiring $s>3$ or producing an atom longer than $10$~s are discarded. These checks establish kinematic and contact feasibility in the fixed-base MuJoCo setting.

\subsubsection{Semantic Verification and Catalog Construction}
Feasibility alone does not ensure that a segment is a meaningful standalone gesture. We therefore first conduct a human motion audit to remove candidates that are incomplete, visually implausible, or lack clear communicative value. Only the approved candidates proceed to semantic annotation.

For each retained atom, we apply Qwen3-VL-8B~\cite{bai2025qwen3} to propose one of the 17 runtime functions in Table~\ref{tab:functions} using the robot-rendered motion, source context, and aligned transcript.  A second human audit confirms or corrects the proposal and removes atoms whose meaning remains ambiguous. Consequently, each catalog entry pairs a feasible trajectory with a human-verified communicative function, and the resulting catalog supplies the approved candidates used by the online Motion Planning Agent.

\begin{algorithm}[t]
\caption{Semantic-Conditioned Motion Planning and State-Aware Execution}
\label{alg:semantic_motion}
\small
\begin{algorithmic}[1]
\Require Motion request $\mathcal{R}_{m}=(f_{1:K},\hat T)$ and
Approved Motion Atom Catalog $\mathcal{C}$
\Ensure Execution of a budget-feasible Atom prefix followed by
neutral recovery

\State $k \gets 1$
\Comment{index of the first unexecuted function}

\While{$k \leq K$}
    \State Measure the current robot state $q$
    \State $B_{\mathrm{rem}} \gets$ current remaining budget
    from $B(\hat{T})$

    \If{$B_{\mathrm{rem}} \leq 0$}
        \State \textbf{break}
    \EndIf

    \State Use dynamic programming to find the largest $L^\star$
    \Statex \hspace{\algorithmicindent}
    satisfying Eq.~(12) for $f_{k:K}$ from state $q$
    within $B_{\mathrm{rem}}$

    \If{$L^\star = 0$}
        \State \textbf{break}
    \EndIf

    \State Use beam search with width 32 to select a feasible
    \Statex \hspace{\algorithmicindent}
    sequence $A_{L^\star}$ for $f_{k:k+L^\star-1}$

    \State Let $a_{i_1}$ be the first Atom in $A_{L^\star}$

    \State Construct $q_b(t)$ from $q$ to
    $q_{i_1}^{\mathrm{in}}$ using Eqs.~(13)--(14)

    \State Prepare the neutral-recovery motion following $a_{i_1}$

    \State Execute the minimum-jerk transition $q_b(t)$
    \State Execute Atom $a_{i_1}$
    \Comment{commit only the first Atom}

    \State Mark $f_k$ as completed
    \State $k \gets k+1$
\EndWhile

\If{$k>1$}
    \State Execute the most recently prepared neutral-recovery motion
\EndIf
\end{algorithmic}
\end{algorithm}
\subsection{Semantic-Conditioned Compositional Planner}
\label{sec:planner}

The online Motion Planning Agent contains the Semantic-Conditioned Compositional Planner and the State-Aware Sequence Executor.

\subsubsection{Semantic Motion Request}

For each spoken reply, the Dialogue Agent produces the response content and the semantic motion request $\mathcal{R}m$ defined in Eq.~(\ref{eq:request}). The ordered functions specify what the body motion should communicate, while atom selection and transition construction remain within the Motion Planning Agent. This separation allows the same reply-level semantics to be grounded according to the available Approved Motion Atoms and the current robot state. The estimated reply duration $\hat T$ is converted into a motion budget
$B(\hat T)$. If the semantic request is received after speech has started, only the remaining budget is used for planning. The function sequence may also be empty, in which case the reply is produced without a body gesture.

\subsubsection{Budget-Constrained Atom Composition}

For each requested function $f_k$, the planner retrieves the approved atoms whose verified label matches $f_k$. Selecting atoms by their individual durations alone is insufficient because an executable sequence must also include the transition from the current robot state, transitions between consecutive atoms, and the final return to the neutral pose.

Let $A_L=(a_{i_1},a_{i_2},\ldots,a_{i_L})$ denote a sequence selected for the first $L$ requested functions. 
Its complete execution cost from the measured state $q$ is defined as
\begin{equation}
\begin{split}
J(A_L;q)
={}& C_{\mathrm{tr}}(q,a_{i_1})
   +\sum_{k=1}^{L}d_{i_k} \\
  &+\sum_{k=1}^{L-1}
    C_{\mathrm{tr}}(a_{i_k},a_{i_{k+1}})
   +C_{\mathrm{ret}}(a_{i_L})+G ,
\end{split}
\label{eq:budget}
\end{equation}
where $d_{i_k}$ is the duration of an Atom,
$C_{\mathrm{tr}}$ denotes transition duration, $C_{\mathrm{ret}}$ denotes neutral-recovery duration, and $G$ is a timing margin. The Semantic-Conditioned Compositional Planner selects the longest feasible action sequence that preserves the requested function order and can be completed within the available budget:
\begin{equation}
\begin{aligned}
\max_{L,A_L}\quad & L\\
\mathrm{s.t.}\quad
    &\phi_{i_k}=f_k,\qquad k=1,\ldots,L,\\
    &J(A_L;q)\leq B(\hat T),
\end{aligned}
\label{eq:prefix}
\end{equation}
where $\phi_{i_k}$ is the audited function label of atom $a_{i_k}$.

\begin{table*}[t!]
\centering
\caption{The 12 dialogue cases used in the end-to-end evaluation and their planned
communicative-function sequences. Arrows indicate execution order, and the
terminal {neutral} state denotes recovery after motion execution.}
\label{tab:planning_cases}

\footnotesize
\setlength{\tabcolsep}{5pt}
\renewcommand{\arraystretch}{1.15}

\begingroup
\begin{tabularx}{\textwidth}{
>{\centering\arraybackslash}p{0.032\textwidth}
>{\raggedright\arraybackslash}p{0.4\textwidth}
>{\raggedright\arraybackslash}X}
\toprule
\textbf{Case}
&
\textbf{User utterance}
&
\textbf{Planned communicative-function sequence}
\\
\midrule

1
&
Introduce artificial intelligence to me.
&
\texttt{turn\_open}
$\rightarrow$ \texttt{emphasize}
$\rightarrow$ \texttt{depict\_size\_shape}
$\rightarrow$ \texttt{neutral}
\\

2
&
Nice to meet you. Bye-bye.
&
\texttt{greet\_farewell}
$\rightarrow$ \texttt{neutral}
\\

3
&
What are three ways to relax?
&
\texttt{enumerate}
$\rightarrow$ \texttt{enumerate}
$\rightarrow$ \texttt{enumerate}
$\rightarrow$ \texttt{neutral}
\\

4
&
Do you like swimming?
&
\texttt{refer\_self}
$\rightarrow$ \texttt{deny}
$\rightarrow$ \texttt{turn\_yield}
$\rightarrow$ \texttt{neutral}
\\

5
&
What is the shape of a basketball?
&
\texttt{depict\_size\_shape}
$\rightarrow$ \texttt{certainty}
$\rightarrow$ \texttt{neutral}
\\

6
&
Which way does the sun move across the sky?
&
\texttt{depict\_direction\_motion}
$\rightarrow$ \texttt{neutral}
\\

7
&
Coffee or tea---which is better?
&
\texttt{contrast\_compare}
$\rightarrow$ \texttt{uncertainty}
$\rightarrow$ \texttt{neutral}
\\

8
&
I really like your idea. It is very helpful.
&
\texttt{thank}
$\rightarrow$ \texttt{neutral}
\\

9
&
I review my notes every evening. Should I continue?
&
\texttt{affirm}
$\rightarrow$ \texttt{emphasize}
$\rightarrow$ \texttt{neutral}
\\

10
&
I am planning a short trip. What should we discuss first?
&
\texttt{enumerate}
$\rightarrow$ \texttt{refer\_external}
$\rightarrow$ \texttt{neutral}
\\

11
&
Let us begin our discussion today.
&
\texttt{turn\_open}
$\rightarrow$ \texttt{neutral}
\\

12
&
My roommate used my laptop without asking, and we argued.
How should I talk to them?
&
\texttt{refer\_self}
$\rightarrow$ \texttt{refer\_other}
$\rightarrow$ \texttt{neutral}
\\

\bottomrule
\end{tabularx}
\endgroup
\end{table*}

The planner first uses dynamic programming to determine the maximum number $L^\star$ of consecutively requested functions that can be realized within the motion budget. It then applies a deterministic beam search with a width of 32 to select an atom sequence for these $L^\star$ functions. Candidate sequences are ranked by smaller unused budget, fewer repetitions of recently executed atoms, and lower imbalance among left-, right-, and bilateral gestures. Because this ranking is performed only among sequences of length $L^\star$, semantic coverage and function order remain unchanged.

\subsubsection{State-aware transitions and execution}

A sequence planned from catalog endpoints alone may become inconsistent with the robot state reached during execution. The State-Aware Sequence Executor
therefore does not commit the complete sequence open loop. Instead, it executes one atom at a time, observes the resulting robot state, and replans the
unexecuted semantic atoms.

Before each Atom, the executor connects the measured state $q_0$ to the atom's initial pose $q_1$ using a minimum-jerk trajectory:
\begin{equation}
\begin{aligned}
q_b(t)
  &=q_0+s(t/T_b)(q_1-q_0),\\
s(u)
  &=10u^3-15u^4+6u^5,\qquad u=t/T_b.
\end{aligned}
\label{eq:bridge}
\end{equation}

By limiting the bridge velocity to $1$~rad/s, its duration satisfies
\begin{equation}
T_b
\geq
1.875
\max_{j\in\mathcal{J}}
|q_{1,j}-q_{0,j}|,
\label{eq:}
\end{equation}
where $1.875$ is the maximum derivative of the normalized minimum-jerk
polynomial and $\mathcal{J}$ denotes the set of controlled joints. This interpolation has zero velocity and acceleration at both endpoints,
providing a smooth connection without modifying the approved atom itself.

After each atom, the executor reads the resulting state and replans the remaining functions under the updated budget. The requested order is fixed, whereas atom selection and transition construction can change at every action boundary. When a visible gesture is required despite insufficient budget, an optional force-first mode executes the lowest-cost atom matching $f_1$, completes neutral recovery, and discards the remaining functions. This mode is handled separately from the budget-feasible solution in Eq.~\eqref{eq:prefix}.

Algorithm~\ref{alg:semantic_motion} summarizes the
semantic-conditioned planning and state-aware execution procedure.
Only the first Atom of each planned sequence is executed; the
remaining functions are replanned from the measured robot state
and updated motion budget.

\subsection{Streaming Face-Speech-Body Integration}
\label{sec:integration}

TRABot integrates the three online agents into a unified real-time interaction loop with a physical G1 humanoid. For each user turn, the Dialogue Agent produces a streaming spoken reply and the semantic motion request $\mathcal{R}_m$. The reply audio provides both the speech output and a shared temporal reference for facial and bodily behavior.

The Facial Animation Agent, implemented with SoulX-FlashHead~\cite{yu2026soulx}, converts the incoming audio stream into facial frames during speech playback. In parallel, the Motion Planning Agent applies the Semantic-Conditioned Compositional Planner to select an ordered sequence of approved motion atoms from $\mathcal{R}_m$. Its State-Aware Sequence Executor constructs transitions from the current robot state and executes the resulting trajectories on the physical G1. Finally, speech playback, facial rendering, and body execution proceed concurrently as a unified embodied response. The shared audio stream aligns facial animation with speech, while the ordered communicative functions and reply-derived motion budget coordinate body motion with the spoken response. This organization preserves the specialized generation mechanism of each modality while coordinating them at the level of a shared dialogue reply, enabling the agent to talk, render facial behavior, and perform compositional gestures during real-time interaction.


\section{Experiments}
\label{sec:experiments}

\subsection{Experimental Setup}
\label{sec:experimental_setup}

\noindent \textbf{Evaluation Metrics.} 
To evaluate the temporal alignment between the generated action sequence and the accompanying audio, we define Binary BAS as the proportion of emphasized speech intervals within which at least one gesture apex occurs. Following the Gaussian nearest-beat formulation of~\cite{li2021aichoreographer} and its adaptation to audio-to-motion generation by~\cite{zhou2023ude}, we compute Gaussian BAS with $\sigma=0.30 \mathrm{s}$. Higher scores indicate stronger temporal alignment between speech emphasis and gesture apexes.

\noindent \textbf{Implementation Details.} 
The online dialogue system uses Gemini 3.1 Flash Live~\cite{team2023gemini}
to generate streaming reply audio together with the ordered communicative functions and estimated reply duration. 
SoulX-FlashHead~\cite{yu2026soulx} generates facial animation from the returned audio, while Qwen3-VL-8B-Instruct~\cite{bai2025qwen3} is used only for offline Motion Atom annotation.
The embodiment follows the 29-DoF G1 joint convention, with 14 upper-body joints controlled for conversational gestures. Executable trajectories retain the original 30-Hz retargeted motion, while body rendering and facial presentation operate at 25~FPS. 
For adaptive activity
detection, we set
$(\kappa_{\mathrm{on}},\kappa_{\mathrm{off}})=(3,1.5)$,
$(\alpha_{\mathrm{on}},\alpha_{\mathrm{off}})=(0.20,0.08)$, and
$(\beta_{\mathrm{on}},\beta_{\mathrm{off}})=(0.55,0.25)$.
Boundary refinement uses
$(\lambda_E,\lambda_A,\lambda_P)=(0.60,0.20,0.20)$ and a 0.32-s search
window. Candidate durations are restricted to $[0.4,4]$~s. Validation uses a
0.05-rad joint-limit margin and velocity, acceleration, and jerk limits of
$7$~rad/s, $120$~rad/s$^2$, and $2200$~rad/s$^3$, respectively. For online
planning, we use $B(\hat T)=3\hat T$, a guard interval of $G=0.35$~s, and a
bridge velocity limit of $1$~rad/s. 

\begin{figure*}[t]
\centering
\includegraphics[width=\textwidth]{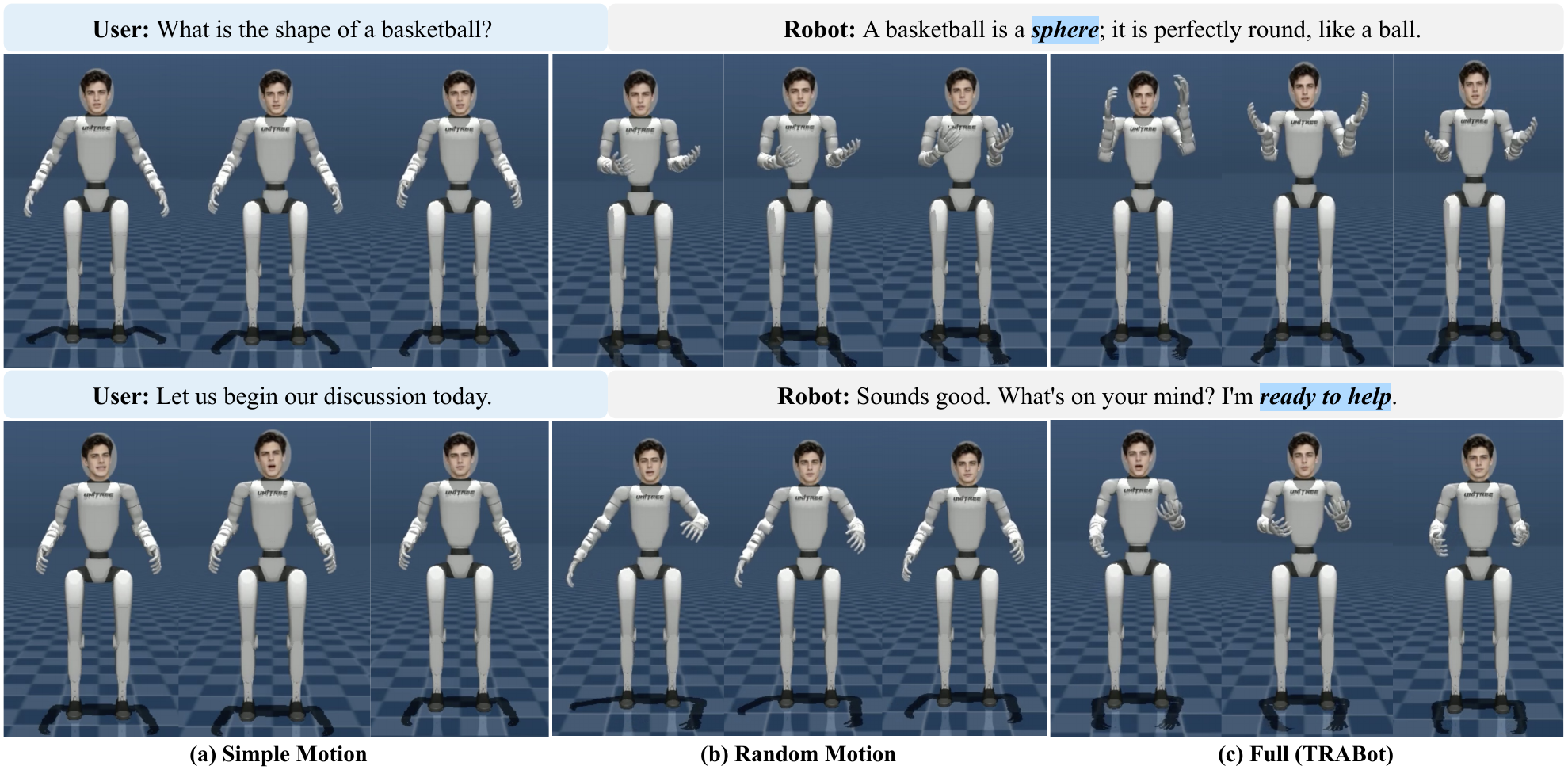}
\caption{Qualitative visualizations of action-speech alignment from the user study. Each row shows a user utterance, the robot's spoken reply, and body-motion keyframes under Simple Motion, Random Motion, and Full (TRABot). Blue-highlighted words or phrases indicate the speech content corresponding to the displayed actions.}
\label{fig:qualitative}
\end{figure*}

\subsection{End-to-End System Evaluation}
\label{sec:e2e_eval}

\paragraph{Overall performance}
We evaluate the complete system over the 12 dialogue turns listed in Table~\ref{tab:planning_cases}, which shows the user utterance and the planned communicative-function sequence for each reply. For each turn, we measure the delays from the onset of reply audio to the first facial frame and the beginning of body motion, the temporal overlap between speech and body motion, and whether all requested modalities complete without failure. All results are reported as average values. As reported in Table~\ref{tab:overall_performance}, speech and body motion overlap for 84.72\% of their execution, and 100\% of dialogue turns complete without failure. These results demonstrate that TRABot reliably coordinates speech, facial animation, and body motion during real-time interaction.

\begin{table}[t!]
\centering
\caption{Overall system performance on 12 utterances.}
\label{tab:overall_performance}
\setlength{\tabcolsep}{6pt}
\begin{tabularx}{\linewidth}{
>{\centering\arraybackslash}p{0.16\linewidth}
>{\centering\arraybackslash}p{0.16\linewidth}
>{\centering\arraybackslash}p{0.19\linewidth}
>{\centering\arraybackslash}X}
\toprule
\makecell[c]{Face offset\\(ms) $\downarrow$} &
\makecell[c]{Body offset\\(ms) $\downarrow$} &
\makecell[c]{Speech--body\\overlap (\%) $\uparrow$} &
\makecell[c]{Completed\\without failure (\%) $\uparrow$} \\
\midrule
37.5
& $-666.6$
& 84.72
& 100 \\
\bottomrule
\end{tabularx}
\end{table}

\begin{table}[t]
\centering
\setlength{\tabcolsep}{14pt}
\caption{Beat alignment score (12 utterances per condition). Binary BAS counts postural apexes inside model-independent emphasis windows; Gaussian BAS uses $\sigma=0.30$ s.}
\label{tab:beat_smoothness_ablation}
\resizebox{1.0\linewidth}{!}{%
\begin{tabular}{lcc}
\toprule
Method & Binary BAS $\uparrow$ & Gaussian BAS $\uparrow$  \\
\midrule
Simple Motion & 0.297 $\pm$ 0.206 & 0.210 $\pm$ 0.129  \\ 
Random Motion & 0.281 $\pm$ 0.197 & 0.190 $\pm$ \textbf{0.145} \\ 
\textbf{Ours (Full model)} & \textbf{0.308} $\pm$ \textbf{0.141} & \textbf{0.225} $\pm$ 0.130  \\
\bottomrule
\end{tabular}}
\label{tab:bas}
\end{table}

\paragraph{Quantitative results}
As shown in Table~\ref{tab:bas}, our full model achieves the highest mean scores for both Binary BAS ($0.308$) and Gaussian BAS ($0.225$). It improves over Simple Motion by $0.011$ and $0.015$, and over Random Motion by $0.027$ and $0.035$, respectively. These consistent gains, together with the lower Binary BAS variability, suggest better temporal alignment between speech emphasis and gesture apexes.

\paragraph{User study}
We conduct a user study with 20 participants to evaluate the perceived benefit
of the complete framework. Five conditions are compared: speech only
({Speech}); speech with audio-driven facial animation
({Speech+Face}); speech and face with body motion randomly composed from
a small set of simple actions
without using our Motion Atom Construction pipeline ({Simple Motion}); speech and face with body motion randomly composed from our approved
motion atoms ({Random Motion}); and the complete system with
semantic-conditioned compositional planner ({Full}). The random condition
uses the same motion catalog and the same amount of body motion
as our full method, but does not use the communicative functions of the reply. Thus, {Simple Motion} evaluates
the benefit of constructing robot-ready Motion Atoms, whereas
{Random Motion} isolates the contribution of semantic-conditioned
motion planning.

\begin{table}[t]
\centering
\caption{
User study results on different conditions (12 utterances per condition). Mean within-turn rank (1 = worst, 5 = best).
}
\label{tab:system_results}
\setlength{\tabcolsep}{2pt}
\resizebox{\linewidth}{!}{%
\begin{tabular}{l|cccccc}
\toprule
Condition &
Natural $\uparrow$ &
Express
$\uparrow$ &
Coherence $\uparrow$ &
Gesture-speech $\uparrow$ &
Preference (\%) $\uparrow$ \\
\midrule
{Speech} &
1.0958& 1.1042& 1.0958 & N/A &0.83 \\

{Speech+Face} &
2.1042& 2.0958&  2.0833& N/A & 0.42\\

{Simple Motion} &
3.1458&3.1333 & 3.1583 & 3.1458 &0.42 \\

{Random Motion} &
3.8667& 3.9083& 3.8708&3.8625 & 13.75\\

Full model&
\textbf{4.7875}&
\textbf{4.7583}&
\textbf{4.7917}&
\textbf{4.7708}&
\textbf{84.58}\\
\bottomrule
\end{tabular}}
\end{table}

\begin{table}[t]
\caption{Evaluation on 200 fixed replies.}
\label{tab:planner_eval}
\centering
\footnotesize
\setlength{\tabcolsep}{4pt}
\begin{tabular}{lcc}
\toprule
Method &
\shortstack{Planned Atoms per reply $\uparrow$} &
\shortstack{Unused budget (s) $\downarrow$} \\
\midrule
Random Match & 1.451 & 2.376 \\
\textbf{Ours (Full model)}  & \textbf{1.705} & \textbf{0.993} \\
\bottomrule
\end{tabular}
\end{table}

In practice, each participant is required to evaluate 12 dialogue turns. For each turn, participants separately rank the five conditions in terms of naturalness, expressiveness, multimodal coherence, and gesture–speech correspondence, and then select their overall preferred condition. 
As reported in Table~\ref{tab:system_results}, adding facial animation changes perceived naturalness/expressiveness from 1.0958/1.1042 to 2.1042/2.0958. Semantic-conditioned planning improves gesture-speech correspondence from 3.8625 under random motion to 4.7708.
Full was selected in 203 of the 240 participant-turn decisions (84.58\%). 
These results further demonstrate the benefit of integrating facial animation with semantic-conditioned motion planning.

\paragraph{Qualitative analysis}
Figure~\ref{fig:qualitative} visualizes representative examples from the user study and compares action--speech alignment across the three conditions. In each example, the blue-highlighted word or phrase in the robot's reply identifies the semantic content to which the displayed body action should correspond. Whereas Simple Motion and Random Motion can produce actions that are weakly related or unrelated to the highlighted speech, the Full TRABot system generates actions that more clearly reflect its meaning. These examples qualitatively illustrate the improved semantic correspondence between spoken content and robot body motion.

\subsection{Automatic Planner Evaluation}

We evaluate planning effectiveness on 200 fixed replies to social-dialogue questions. The reply, ordered functions, and estimated duration $\hat{T}$ are frozen before evaluation. Both planners then run fully offline using the same Motion Atom Catalog and budget $B=3\hat{T}$, including transition, neutral-recovery, and guard costs.
{Random Match} greedily samples a feasible function-matched atom and is averaged over 50 trials per reply. {Full model} searches for the longest feasible ordered-function prefix. As shown in Table~\ref{tab:planner_eval}, {Full} model increases the mean number of planned Motion Atoms per reply by 17.5\% and reduces unused budget by 58.2\% relative to {Random Match}, demonstrating higher communicative-function coverage and more efficient budget utilization under the same motion constraints.

\section{Conclusion}
We presented TRABot, an agent-based framework for real-time
face-speech-body interaction on a humanoid robot. First, TRABot constructs a catalog of robot-ready Motion Atoms with validated trajectories and human-audited communicative functions. Second, its semantic-conditioned planner
composes these atoms according to the communicative intent and duration of each reply while accounting for transition costs and the current robot state. Third, the framework integrates the planned body motions with streaming speech and audio-driven facial animation to support real-time multimodal dialogue. The complete system achieved the strongest beat-alignment scores and received the highest rankings across all user-study criteria. These results demonstrate that semantic motion composition provides an effective interface between conversational intent and executable multimodal robot behavior. 
A current limitation is that the diversity of generated body behavior is bounded by the coverage of the existing Motion Atom catalog.

\section*{Acknowledgments}
Illustrations in Fig.~\ref{fig:overview} were generated by ChatGPT.

{\small
\bibliographystyle{IEEEtran}
\bibliography{refs}}

@String(ECCV= {Eur. Conf. Comput. Vis.})

@INPROCEEDINGS{peng2024synctalk,
  author={Peng, Ziqiao and Hu, Wentao and Shi, Yue and Zhu, Xiangyu and Zhang, Xiaomei and Zhao, Hao and He, Jun and Liu, Hongyan and Fan, Zhaoxin},
  booktitle={Proceedings of the IEEE/CVF Conference on
               Computer Vision and Pattern Recognition}, 
  title={SyncTalk: The Devil is in the Synchronization for Talking Head Synthesis}, 
  year={2024},
  volume={},
  number={},
  pages={666-676},
}

@inproceedings{tian2024emo,
  author    = {Tian, Linrui and Wang, Qi and Zhang, Bang and Bo, Liefeng},
  title     = {{EMO}: Emote Portrait Alive---Generating Expressive Portrait Videos with {Audio2Video} Diffusion Model under Weak Conditions},
  booktitle = {Computer Vision -- ECCV 2024},
  series    = {Lecture Notes in Computer Science},
  volume    = {15141},
  pages     = {244--260},
  year      = {2024}
}

@inproceedings{ki2025float,
  author={Ki, Taekyung and Min, Dongchan and Chae, Gyeongsu},
  booktitle={Proceedings of the IEEE/CVF International
               Conference on Computer Vision}, 
  title={FLOAT: Generative Motion Latent Flow Matching for Audio-Driven Talking Portrait}, 
  year={2025},
  volume={},
  number={},
  pages={14699-14710},
 }

@inproceedings{zhen2025teller,
  author={Zhen, Dingcheng and Yin, Shunshun and Qin, Shiyang and Yi, Hou and Zhang, Ziwei and Liu, Siyuan and Qi, Gan and Tao, Ming},
  booktitle={Proceedings of the IEEE/CVF Conference on
               Computer Vision and Pattern Recognition}, 
  title={Teller: Real-Time Streaming Audio-Driven Portrait Animation with Autoregressive Motion Generation}, 
  year={2025},
  volume={},
  number={},
  pages={21075-21085},
 }

@inproceedings{cassell2001beat,
  author    = {Cassell, Justine and Vilhj{\'a}lmsson, Hannes H{\"o}gni and Bickmore, Timothy},
  title     = {{BEAT}: the Behavior Expression Animation Toolkit},
  booktitle = {Proceedings of the 28th Annual Conference on Computer Graphics and Interactive Techniques},
  year      = {2001},
  pages     = {477--486},
}

@INPROCEEDINGS{ConvoFusion2024,
  author={Mughal, Muhammad Hamza and Dabral, Rishabh and Habibie, Ikhsanul and Donatelli, Lucia and Habermann, Marc and Theobalt, Christian},
  booktitle={Proceedings of the IEEE/CVF Conference on Computer Vision and Pattern Recognition}, 
  title={ConvoFusion: Multi-Modal Conversational Diffusion for Co-Speech Gesture Synthesis}, 
  year={2024},
  volume={},
  number={},
  pages={1388-1398},}

@INPROCEEDINGS{HOP2025,
  author={Cheng, Hongye and Wang, Tianyu and Shi, Guangsi and Zhao, Zexing and Fu, Yanwei},
  booktitle={Proceedings of the IEEE/CVF Conference on Computer Vision and Pattern Recognition}, 
  title={HOP: Heterogeneous Topology-based Multimodal Entanglement for Co-Speech Gesture Generation}, 
  year={2025},
  volume={},
  number={},
  pages={906-916},
 }

@INPROCEEDINGS{GestureHYDRA2025,
  author={Yang, Quanwei and Huang, Luying and Wang, Kaisiyuan and Guan, Jiazhi and He, Shengyi and Li, Fengguo and Zhou, Hang and Yu, Lingyun and Li, Yingying and Feng, Haocheng and Xie, Hongtao},
  booktitle={Proceedings of the IEEE/CVF International Conference on Computer Vision}, 
  title={GestureHYDRA: Semantic Co-Speech Gesture Synthesis via Hybrid Modality Diffusion Transformer and Cascaded-Synchronized Retrieval-Augmented Generation}, 
  year={2025},
  volume={},
  number={},
  pages={12615-12625},
  }

@INPROCEEDINGS{SemGes2025,
  author={Liu, Lanmiao and Ghaleb, Esam and Özyurek, Asli and Yumak, Zerrin},
  booktitle={Proceedings of the IEEE/CVF International Conference on Computer Vision}, 
  title={SemGes: Semantics-Aware Co-Speech Gesture Generation Using Semantic Coherence and Relevance Learning}, 
  year={2025},
  volume={},
  number={},
  pages={13963-13973},
  }

@inproceedings{yoon2019robots,
  author    = {Yoon, Youngwoo and Ko, Woo-Ri and Jang, Minsu and Lee, Jaeyeon and Kim, Jaehong and Lee, Geehyuk},
  title     = {Robots Learn Social Skills: End-to-End Learning of Co-Speech Gesture Generation for Humanoid Robots},
  booktitle = {International Conference on Robotics and Automation},
  year      = {2019},
  pages     = {4303--4309},
}

@inproceedings{kucherenko2020gesticulator,
  author    = {Kucherenko, Taras and Jonell, Patrik and van Waveren, Sanne and Henter, Gustav Eje and Alexanderson, Simon and Leite, Iolanda and Kjellstr{\"o}m, Hedvig},
  title     = {Gesticulator: A Framework for Semantically-Aware Speech-Driven Gesture Generation},
  booktitle = {Proceedings of the 2020 International Conference on Multimodal Interaction},
  year      = {2020},
  pages     = {242--250},
}

@article{yoon2020trimodal,
  author    = {Yoon, Youngwoo and Cha, Bok and Lee, Joo-Haeng and Jang, Minsu and Lee, Jaeyeon and Kim, Jaehong and Lee, Geehyuk},
  title     = {Speech Gesture Generation from the Trimodal Context of Text, Audio, and Speaker Identity},
  journal   = {ACM Transactions on Graphics},
  year      = {2020},
  volume    = {39},
  number    = {6},
  articleno = {222},
  pages     = {222:1--222:16},
}

@inproceedings{lu2023cospeech,
  author    = {Lu, Shuhong and Yoon, Youngwoo and Feng, Andrew},
  title     = {Co-Speech Gesture Synthesis using Discrete Gesture Token Learning},
  booktitle = {IEEE/RSJ International Conference on Intelligent Robots and Systems},
  year      = {2023},
  pages     = {9808--9815},
}

@inproceedings{nguyen2023gesticulating,
  author    = {Nguyen, Tan Viet Tuyen and Schmuck, Viktor and {\c{C}}eliktutan, Oya},
  title     = {Gesticulating with {NAO}: Real-Time Context-Aware Co-Speech Gesture Generation for Human-Robot Interaction},
  booktitle = {Companion Publication of the 25th International Conference on Multimodal Interaction},
  year      = {2023},
  pages     = {112--114},
}

@inproceedings{richard2021meshtalk,
  author    = {Richard, Alexander and Zollh{\"o}fer, Michael and Wen, Yandong and de la Torre, Fernando and Sheikh, Yaser},
  title     = {{MeshTalk}: {{3D}} Face Animation From Speech Using Cross-Modality Disentanglement},
  booktitle = {Proceedings of the IEEE/CVF International Conference on Computer Vision},
  pages     = {1173--1182},
  year      = {2021},
}

@inproceedings{fan2022faceformer,
  author    = {Fan, Yingruo and Lin, Zhaojiang and Saito, Jun and Wang, Wenping and Komura, Taku},
  title     = {{FaceFormer}: Speech-Driven {{3D}} Facial Animation With Transformers},
  booktitle = {Proceedings of the IEEE/CVF Conference on Computer Vision and Pattern Recognition},
  pages     = {18770--18780},
  year      = {2022}
}

@inproceedings{xing2023codetalker,
  author    = {Xing, Jinbo and Xia, Menghan and Zhang, Yuechen and Cun, Xiaodong and Wang, Jue and Wong, Tien-Tsin},
  title     = {{CodeTalker}: Speech-Driven {{3D}} Facial Animation With Discrete Motion Prior},
  booktitle = {Proceedings of the IEEE/CVF Conference on Computer Vision and Pattern Recognition},
  pages     = {12780--12790},
  year      = {2023}
}

@inproceedings{liuhaiyang2024emage,
  title={Emage: Towards unified holistic co-speech gesture generation via expressive masked audio gesture modeling},
  author={Liu, Haiyang and Zhu, Zihao and Becherini, Giorgio and Peng, Yichen and Su, Mingyang and Zhou, You and Zhe, Xuefei and Iwamoto, Naoya and Zheng, Bo and Black, Michael J},
  booktitle={IEEE/CVF Conference on Computer Vision and Pattern Recognition},
  pages={1144--1154},
  year={2024},
}

@inproceedings{chen2024diffsheg,
  author    = {Chen, Junming and Liu, Yunfei and Wang, Jianan and Zeng, Ailing and Li, Yu and Chen, Qifeng},
  title     = {{DiffSHEG}: A Diffusion-Based Approach for Real-Time Speech-driven Holistic {{3D}} Expression and Gesture Generation},
  booktitle = {Proceedings of the IEEE/CVF Conference on Computer Vision and Pattern Recognition},
  pages     = {7352--7361},
  year      = {2024},
}

@InProceedings{Mughal2026CVPR,
    author    = {Mughal, M. Hamza and Dabral, Rishabh and Demberg, Vera and Theobalt, Christian},
    title     = {{MIBURI: Towards Expressive Interactive Gesture Synthesis}},
    booktitle = {IEEE/CVF Conference on Computer Vision and Pattern Recognition},
    year      = {2026},
    pages     = {40031-40041}
}

@inproceedings{hogue2025joint,
  author    = {Hogue, Steven and Zhang, Chenxu and Tian, Yapeng and Guo, Xiaohu},
  title     = {Joint Co-Speech Gesture and Expressive Talking Face Generation using Diffusion with Adapters},
  booktitle = {Proceedings of the IEEE/CVF Winter Conference on Applications of Computer Vision},
  pages     = {4163--4172},
  year      = {2025},
}

@inproceedings{gomez2020holistic,
  author    = {Gomez, Randy and Szapiro, Deborah and Merino, Luis and Nakamura, Keisuke},
  title     = {A Holistic Approach in Designing Tabletop Robot's Expressivity},
  booktitle = {IEEE International Conference on Robotics and Automation},
  year      = {2020},
  pages     = {1970--1976},
}

@inproceedings{cherakara2023furchat,
  author    = {Cherakara, Neeraj and Varghese, Finny and Shabana, Sheena and Nelson, Nivan and others},
  title     = {{FurChat}: An Embodied Conversational Agent using {LLM}s, Combining Open and Closed-Domain Dialogue with Facial Expressions},
  booktitle = {the Special Interest Group on Discourse and Dialogue},
  year      = {2023},
  pages     = {588--592}
}

@inproceedings{wang2024misbehavin,
  author    = {Wang, Zining and Reisert, Paul and Nichols, Eric and Gomez, Randy},
  title     = {Ain't Misbehavin'---Using {LLM}s to Generate Expressive Robot Behavior in Conversations with the Tabletop Robot {Haru}},
  booktitle = {Companion of the ACM/IEEE International Conference on Human-Robot Interaction},
  year      = {2024},
  pages     = {1105--1109}
}

@inproceedings{mahadevan2024genem,
  author    = {Mahadevan, Karthik and Chien, Jonathan and Brown, Noah and Xu, Zhuo and Parada, Carolina and others},
  title     = {Generative Expressive Robot Behaviors using Large Language Models},
  booktitle = {Proceedings of the ACM/IEEE International Conference on Human-Robot Interaction},
  year      = {2024},
  pages     = {482--491},
}

@article{galatolo2025simultaneous,
  author  = {Galatolo, Alessio and Winkle, Katie},
  title   = {Simultaneous Text and Gesture Generation for Social Robots with Small Language Models},
  journal = {Frontiers in Robotics and AI},
  year    = {2025},
  volume  = {12},
  pages   = {1581024}
}

@inproceedings{abbo2026expressive,
  author    = {Abbo, Giulio Antonio and Janssens, Ruben and {Van de Vreken}, Seppe and Belpaeme, Tony},
  title     = {Expressive {Furhat}: Generating Real-Time Facial Expressions for Human-Robot Dialogue with {LLM}s},
  booktitle = {ACM/IEEE International Conference on Human-Robot Interaction},
  year      = {2026},
  pages     = {456--460},
}

@inproceedings{zhou2023ude,
  author    = {Zhou, Zixiang and Wang, Baoyuan},
  title     = {{UDE}: A Unified Driving Engine for Human Motion Generation},
  booktitle = {Proceedings of the IEEE/CVF Conference on Computer Vision and Pattern Recognition},
  pages     = {5632--5641},
  year      = {2023}
}

@inproceedings{li2021aichoreographer,
  author    = {Li, Ruilong and Yang, Shan and Ross, David A. and
               Kanazawa, Angjoo},
  title     = {AI Choreographer: Music Conditioned 3D Dance Generation With AIST++},
  booktitle = {Proceedings of the IEEE/CVF International Conference on Computer Vision},
  pages     = {13401--13412},
  year      = {2021}
}

@inproceedings{nichols2026holistic,
  author    = {Nichols, Eric and {Mejia Tobar}, Miguel and Gomez, Randy},
  title     = {Holistic {LLM}-Based Expressive Behavior Generation for Robots},
  booktitle = {Companion Proceedings of the 21st ACM/IEEE International Conference on Human-Robot Interaction},
  year      = {2026},
  pages     = {809--813},
}

@article{yu2026soulx,
  title={SoulX-FlashHead: Oracle-guided Generation of Infinite Real-time Streaming Talking Heads},
  author={Yu, Tan and Qiao, Qian and Shen, Le and Zhou, Ke and Hu, Jincheng and others},
  journal={arXiv preprint arXiv:2602.07449},
  year={2026}
}

@inproceedings{wang2026robogesture,
  title={{RoboGesture: Real-Time Semantic-aligned Co-Speech Gestures Generation for Humanoid Interaction}},
  author={Wang, Zifan and Ren, Ziang and Shi, Pengyang and Wang, Zirui and Lin, Chenghuai
         and Wang, Tianze and Qi, Zekun and Zhao, Liangliang and Wang, He and Yi, Li},
  booktitle={European Conference on Computer Vision},
  year={2026}
}

@inproceedings{huang2025liveavatar,
      title={{Live Avatar: Streaming Real-time Audio-Driven Avatar Generation with Infinite Length}}, 
      author={Yubo Huang and Hailong Guo and Fangtai Wu and Shifeng Zhang and Shijie Huang and Qijun Gan and Lin Liu and Sirui Zhao and Enhong Chen and Jiaming Liu and Steven Hoi},
       booktitle={European Conference on Computer Vision},
  year={2026}
}

@article{team2023gemini,
  title={Gemini: a family of highly capable multimodal models},
  author={Team, Gemini and Anil, Rohan and others},
  journal={arXiv preprint arXiv:2312.11805},
  year={2023}
}

@article{bai2025qwen3,
  title={Qwen3-vl technical report},
  author={Bai, Shuai and Cai, Yuxuan and Chen, Ruizhe and Chen, Keqin and Chen, Xionghui and Cheng, Zesen and Deng, Lianghao and Ding, Wei and Gao, Chang and Ge, Chunjiang and others},
  journal={arXiv preprint arXiv:2511.21631},
  year={2025}
}

@article{CommunicativeRobotGesture2012,
  title={Generation and Evaluation of Communicative Robot Gesture},
  author={Maha Salem and Stefan Kopp and Ipke Wachsmuth and Katharina Rohlfing and Frank Joublin},
  journal={International Journal of Social Robotics},
  year={2012},
  Volume={4}, 
  pages={201–217} 
}

\end{document}